\documentclass[sigconf]{acmart}

\usepackage{algorithm}
\usepackage{algpseudocode}
\usepackage{multirow}
\usepackage{amsmath}

\usepackage{amssymb}
\usepackage{amsthm}
\usepackage{graphicx}
\usepackage{enumitem}
\usepackage[table]{xcolor}
\usepackage{makecell}
\theoremstyle{definition}

\renewcommand\footnotetextcopyrightpermission[1]{}

\author{Jseen Zhang}
\authornote{Both authors contributed equally to this research.}
\affiliation{%
  \institution{University of California, San Diego}
  \country{USA} 
}

\author{Gabriel Adineera} 
\authornote{Both authors contributed equally to this research.}
\affiliation{%
  \institution{Texas A\&M University-Commerce}
  \country{USA} 
}

\author{Jinzhou Tan}
\affiliation{%
  \institution{University of California, San Diego} 
  \country{USA} 
}

\author{Jinoh Kim}
\authornote{Corresponding author.} 
\affiliation{%
  \institution{Texas A\&M University-Commerce}
  \country{USA} 
}

\title{ResWM: Residual-Action World Model for Visual RL}
\begin{document}
\title{ResWM: Residual-Action World Model for Visual RL}

\begin{abstract}
Learning predictive world models from raw visual observations is a central challenge in reinforcement learning (RL), especially for robotics and continuous control. Conventional model-based RL frameworks directly condition future predictions on absolute actions, which makes optimization unstable: the optimal action distributions are task-dependent, unknown a priori, and often lead to oscillatory or inefficient control. To address this, we introduce the Residual-Action World Model (ResWM), a new framework that reformulates the control variable from absolute actions to residual actions—incremental adjustments relative to the previous step. This design aligns with the inherent smoothness of real-world control, reduces the effective search space, and stabilizes long-horizon planning.
To further strengthen the representation, we propose an Observation Difference Encoder that explicitly models the changes between adjacent frames, yielding compact latent dynamics that are naturally coupled with residual actions. ResWM is integrated into a Dreamer-style latent dynamics model with minimal modifications and no extra hyperparameters. Both imagination rollouts and policy optimization are conducted in the residual-action space, enabling smoother exploration, lower control variance, and more reliable planning.
Empirical results on the DeepMind Control Suite demonstrate that ResWM achieves consistent improvements in sample efficiency, asymptotic returns, and control smoothness, significantly surpassing strong baselines such as Dreamer and TD-MPC. Beyond performance, ResWM produces more stable and energy-efficient action trajectories, a property critical for robotic systems deployed in real-world environments. These findings suggest that residual action modeling provides a simple yet powerful principle for bridging algorithmic advances in RL with the practical requirements of robotics.
\end{abstract}

\maketitle
\section{Introduction}

Learning world models from high-dimensional visual observations stands as a cornerstone challenge in reinforcement learning (RL)\cite{WM,hafner2019learninglatentdynamicsplanning,silver2016alphago,KABB,GAM,MAB,MAT,VLMDONG,FDWR}, demanding the orchestration of representation learning, dynamics prediction, and policy optimization in tandem\cite{luo2022surveymodelbasedreinforcementlearning,seo2023maskedworldmodelsvisual,3DAgent,CF,MMCOT,HTC,KAF}. In contrast to state-based paradigms where low-dimensional inputs transparently expose underlying system dynamics, visual RL grapples with multifaceted complexity, especially in robotics where agents navigate long-horizon tasks amid perpetually shifting environments\cite{moerland2022modelbasedreinforcementlearningsurvey,DrDiff,OSC}. Although model-free approaches such as SAC\cite{sac} and PPO\cite{ppo} have garnered impressive achievements, their inherent sample inefficiency severely limits their viability in real-world scenarios, where data acquisition is costly, limited, or fraught with risks. Model-based reinforcement learning (MBRL)\cite{moerland2022modelbasedreinforcementlearningsurvey,saanum2024simplifyinglatentdynamicssoftly} emerges as a promising antidote, cultivating an internal world model that facilitates efficient planning and policy refinement via imaginative simulations\cite{DBLP:journals/corr/abs-1708-02596, 10.1145/122344.122377}. Yet, the traditional manner of embedding actions within these models has surfaced as a pivotal bottleneck, hindering the full realization of MBRL's potential.

Predominant world models condition their latent dynamics directly on \emph{absolute actions}, a seemingly intuitive choice that nonetheless instills suboptimal inductive biases with far-reaching implications\cite{ding2025understandingworldpredictingfuture,RAN,POT,SGN,z22}. Primarily, this setup casts policy learning as a high-variance conundrum, compounded by the task-specific, non-stationary distributions of optimal absolute actions. Furthermore, it frequently engenders oscillatory or erratic control trajectories, undermining planning efficacy and introducing safety hazards in physical embodiments. These shortcomings underscore a profound disconnect between the algorithmic scaffolding of world models and the imperatives of smooth, resilient control in embodied agents.

In this paper, we champion the notion that the smoothness intrinsic to continuous control transcends mere desirability, evolving into a foundational principle ripe for exploitation. Our pioneering insight revolves around the \emph{residual change} between successive actions—a quantity far more predictable and tractable than its absolute counterpart. Leveraging this, we unveil the \textbf{Residual-Action World Model (ResWM)}, a transformative framework that reparameterizes the control variable from absolute actions to \emph{residual actions}. This reformulation embeds a robust \textbf{temporal smoothness prior} into the action space, fundamentally curtailing learning complexity by modeling incremental refinements rather than outright commands. In doing so, ResWM harmonizes with the continuum of physical dynamics, furnishing a steadfast bedrock for extended-horizon planning and fostering control signals that are inherently stable and energy-efficient.

To anchor the control signal in the most salient perceptual insights, we complement our residual action paradigm with an innovative \textbf{Observation Difference Encoder (ODL)}. Diverging from the norm of independently encoding static frames, the ODL meticulously distills the embedded dynamics from the disparities between adjacent observations \cite{sun2024learninglatentdynamicrobust,z13,z14,z15,z20}. This yields a compact, \textbf{dynamics-aware latent representation} that seamlessly aligns with the prediction of residual actions, effectively sieving out static redundancies to spotlight temporal shifts causally pivotal to the agent's adaptive adjustments. By prioritizing these differential cues, the ODL imbues the model with a sharpened focus on action-induced changes, enhancing its causal reasoning capabilities.

ResWM is meticulously engineered for effortless assimilation into established Dreamer-style architectures, entailing only minimal alterations and eschewing any novel hyperparameters. Imagination-driven planning and policy optimization unfold wholly within the residual-action domain, transmuting exploration into fluid, localized perturbations rather than volatile, high-variance explorations. This paradigm shift not only bolsters learning stability and sample efficiency but also yields action trajectories that are demonstrably smoother and more energy-conserving—indispensable attributes for the secure integration of robotic systems in real-world contexts.

We substantiate ResWM's efficacy on the \emph{DeepMind Control Suite}, a benchmark par excellence for continuous visual control. Empirical evaluations reveal ResWM's consistent supremacy in sample efficiency, asymptotic performance, and control smoothness, markedly eclipsing formidable baselines such as Dreamer and TD-MPC. Complementary qualitative scrutiny discloses ResWM's propensity for stabler latent dynamics and mitigated compounding prediction errors, affirming its robustness and versatility.

In summary, this work advances the following contributions:
\begin{itemize}
    \item We introduce the \textbf{Residual-Action World Model (ResWM)}, a novel MBRL framework that reparameterizes the action space to instill a powerful smoothness prior, dramatically alleviating learning complexity and augmenting stability.
    \item We propose the \textbf{Observation Difference Encoder}, an architecture that forges a dynamics-aware latent representation by explicitly capturing temporal changes in consecutive visual inputs.
    \item We empirically validate that ResWM delivers substantial performance enhancements across intricate control tasks, providing a principled conduit to reconcile deep RL advancements with the exigencies of real-world robotics.
\end{itemize}

\section{RELATED WORKS}
Model-based reinforcement learning (MBRL) tackles visual RL by learning world models from pixel observations, offering efficiency over model-free approaches like SAC and PPO. Dreamer~\cite{hafner2019dream, hafner2023mastering,M10,M11,M2,M3,M4} employs recurrent state-space models (RSSMs) for latent dynamics, while TD-MPC~\cite{hansen2022temporal,M5} optimizes trajectories in latent space. Recent advances, such as DeepRAD~\cite{guo2023deep} with robust representations and RAD~\cite{laskin2020reinforcement} with data augmentation, condition dynamics on absolute actions. This approach introduces high-variance policies and oscillatory control, limiting stability in robotic tasks—where ResWM instead uses residual actions for smoother, more stable optimization.

To address action instability, residual policies~\cite{silver2018residual} predict incremental adjustments, as seen in incremental learning~\cite{ravi2018learning}, reducing variance in planning. Robotics methods like Gaussian processes~\cite{calandra2016manifold} and filtering~\cite{deisenroth2011pilco} enforce smoothness for energy efficiency. ResAct~\cite{resact2025} applies residuals in actor-critic setups, improving stability on DMControl (e.g., 715 vs. 690 for Quadruped, Walk at 1M steps, Table III), but lacks world model integration for imagination-based learning. ResWM uniquely embeds residuals into RSSMs, enabling latent rollouts and outperforming ResAct (e.g., 644.8 vs. 630.2 average on hard tasks at 1M steps) with a design tailored for long-horizon planning.

For perception, contrastive methods like CURL~\cite{yeh2020curl} and SVEA~\cite{guo2021symbolic} use augmentations, while delta-based approaches~\cite{zhang2020deep} model frame differences. TACO~\cite{taco2023} and MaDi~\cite{madi2024} enhance efficiency but rely on absolute actions, lagging behind ResWM’s 925.0 average on DMControl (Table II) and 0.96 normalized mean on Atari (Table IV). ResWM’s Observation Difference Encoder (ODL) differs by conditioning residuals on explicit observation deltas, creating a causal perception-control link in Dreamer-style models—surpassing TACO (887.1) and MaDi (885.1) while excelling in both continuous and discrete domains.
\section{METHODOLOGY}
\label{sec:method}

Our methodology pioneers a principled reformulation of action and observation representations in latent variable world models, challenging the conventional paradigm of modeling absolute and temporally independent actions. Drawing inspiration from the inherent continuity of physical systems and biological motor control \cite{Banerjee_2025, peters2008reinforcement}, we propose a novel framework anchored on two groundbreaking principles: (1) redefining the control variable as a \textbf{residual action} to embed a robust temporal smoothness prior, thereby transforming chaotic global searches in the action space into elegant, localized refinements, and (2) conditioning this control signal on an explicit encoding of \textbf{observation differences} to forge a highly dynamics-aware latent space that captures the essence of environmental evolution. This innovative synthesis culminates in the \textbf{Residual-Action World Model (ResWM)}, a framework that not only bridges theoretical elegance with practical efficacy but also unlocks unprecedented stability and sample efficiency in visual reinforcement learning (RL). We elaborate on its architectural components, objective functions, and optimization processes below.

\subsection{Preliminaries: Latent Dynamics Models}

We formulate the visual control problem as a Partially Observable Markov Decision Process (POMDP) \cite{sutton2018reinforcement}, formally defined by the tuple $(\mathcal{O}, \mathcal{A}, P, R, \gamma)$. At each discrete time step $t$, the agent receives a high-dimensional visual observation $o_t \in \mathcal{O}$ and emits an action $a_t \in \mathcal{A}$. The environment subsequently transitions to a new unobserved true state according to the transition dynamics $P$, and the agent receives a scalar reward $r_t = R(s_t, a_t)$. The agent's overarching objective is to learn a policy $\pi(a_t|o_{\le t})$ that maximizes the expected discounted return $\mathbb{E}[\sum_{t=0}^T \gamma^t r_t]$, where $\gamma \in [0, 1)$ is the discount factor.

Model-based RL agents approach this intractable high-dimensional problem by learning a generative world model from a history of interactions $\mathcal{D} = \{(o_t, a_t, r_t)\}_{t=0}^T$ \cite{ha2018worldmodels, hafner2019dreamer}. This model typically factorizes into several key components:
\begin{itemize}
    \item \textbf{Representation Model (Encoder)} $h_\theta: o_t \to s_t$, which maps high-dimensional observations to a compact, Markovian latent state space.
    \item \textbf{Transition Model (Dynamics)} $g_\phi(s_{t+1}|s_t, a_t)$, which predicts the forward evolution of the environment purely in the latent space.
    \item \textbf{Reward Predictor} $r_\psi(r_t|s_t, a_t)$, which estimates the immediate task reward to facilitate offline planning or policy optimization \cite{hafner2020mastering}.
\end{itemize}
A fundamental challenge in this paradigm lies in the choice of representation for the action $a_t$. Traditional world models \cite{hafner2023mastering} assume that the optimal action distribution at time $t$ is independent of $a_{t-1}$, which overlooks the non-stationary nature of action distributions and the physical constraints of actuation. This frequently leads to high-frequency action chattering and optimization instability—issues that our novel residual formulation is specifically designed to circumvent.

\begin{figure*}[t]
    \centering
    \resizebox{\textwidth}{!}{%
        \includegraphics{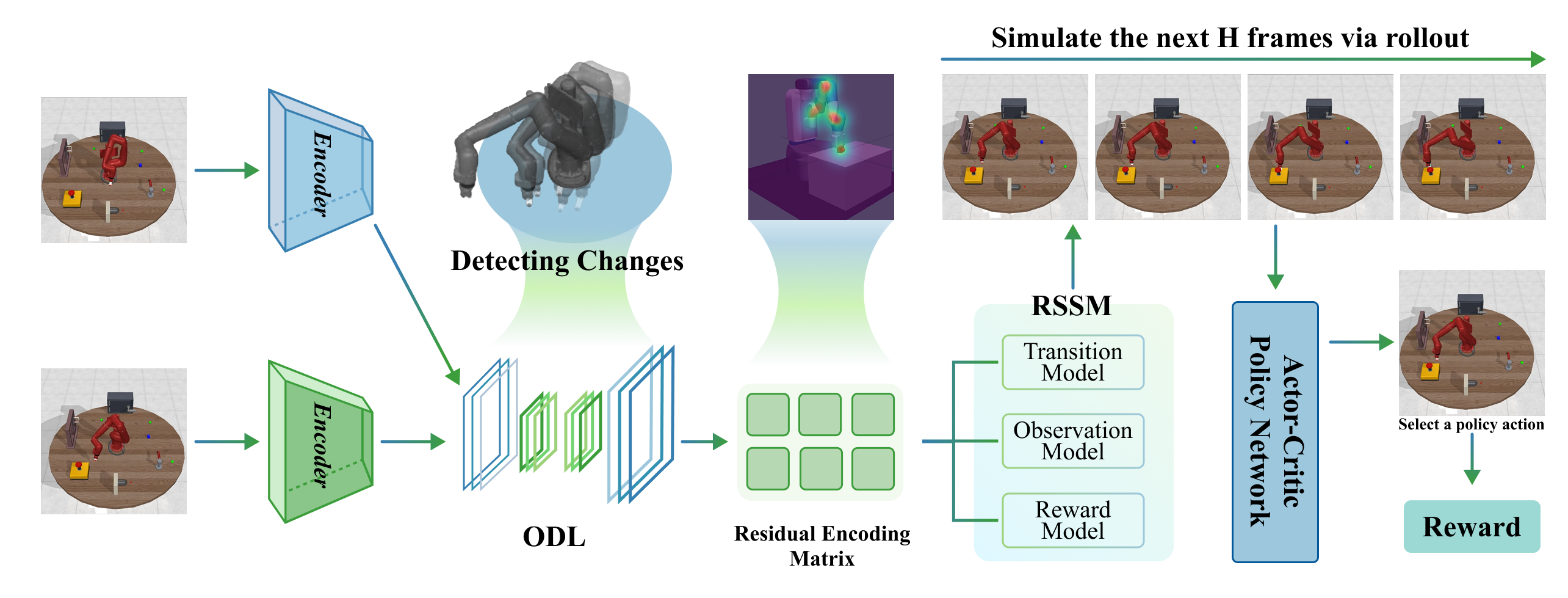}
    }
    \caption{Architecture of the proposed Residual-Action World Model (ResWM). (1) \textbf{Observation Difference Encoder (ODL)}: Consecutive frames $o_{t-1}$ and $o_t$ are processed to extract dynamic deltas, producing a dynamics-aware latent vector $z_t$. (2) \textbf{Residual Policy}: The actor network predicts a residual update $\delta a_t$ conditioned on $z_t$ and the previous action $a_{t-1}$, enforcing temporal smoothness. (3) \textbf{Latent Dynamics}: A Recurrent State-Space Model (RSSM) rolls out future latent states $s_{t+1}$ driven by these residual actions, enabling stable, long-horizon imagination for actor-critic optimization.}
    \label{fig:pipeline}
\end{figure*}

\subsection{The Residual-Action Policy as a Smoothness Prior}

Our central hypothesis posits that the direct prediction of absolute actions $a_t$ constitutes an inherently ill-posed problem for continuous control tasks. Optimal physical trajectories are rarely composed of disjointed, independent control signals; rather, they demand smooth, continuous transitions. To seamlessly integrate this inductive bias, we introduce a reparameterization technique: the policy predicts an incremental adjustment $\delta a_t$ relative to the previous action $a_{t-1}$, effectively anchoring decisions in temporal continuity. The final action emitted to the environment emerges through a composed transformation:
\begin{equation}
a_t = \tanh\!\big(a_{t-1} + \delta a_t\big), \quad \text{where} \quad \delta a_t \sim \pi_\theta(\cdot \mid z_t, a_{t-1}).
\label{eq:residual_action}
\end{equation}
This elegant formulation instills a strong \textbf{temporal smoothness prior} into the policy network. By restricting the control output to a differential term, we reorient the optimization landscape from an expansive, unconstrained search across the global action space $\mathcal{A}$ to a refined exploration within a localized manifold centered on $a_{t-1}$. Consequently, this drastically narrows the effective search space, enhances the sample efficiency of the actor-critic learning phase, and fosters control signals characterized by low-frequency, energy-efficient profiles. Such qualities are paramount for deploying RL algorithms on real-world physical systems, such as robotic manipulators and autonomous vehicles, where mechanical wear and actuation energy are critical constraints. In Equation \ref{eq:residual_action}, $z_t$ serves as a specialized latent variable encapsulating environmental changes, which leads to our next architectural innovation.

\subsection{Observation Difference Encoding for a Dynamics-Aware State}

To equip the policy with the most pertinent perceptual cues necessary for crafting the next action residual, we propose the \textbf{Observation Difference Encoder (ODL)}. Traditional frame-stacking \cite{mnih2015human} implicitly models velocity but often suffers from visual aliasing and high redundancy. Instead, ODL reimagines representation learning by explicitly focusing on temporal deltas. This is grounded in the insight that the optimal residual $\delta a_t$ is predominantly driven by the differential changes between consecutive observations rather than static visual snapshots. We formalize this through a sophisticated mapping $\Phi_{\text{ODL}}$:
\begin{equation}
z_t = \Phi_{\text{ODL}}(o_t, o_{t-1}) = \text{LN}\!\big(\text{FC}(f(o_t) - f'(o_{t-1}))\big),
\label{eq:odl}
\end{equation}
where $f$ and $f'$ denote independent or Siamese Convolutional Neural Network (CNN) encoders, FC is a fully connected layer, and LN denotes Layer Normalization \cite{ba2016layer} applied for representation stability. 

This architectural innovation achieves dual objectives: (1) it functions as a precise temporal filter, distilling dynamic, task-relevant elements (e.g., a moving object or a swinging pendulum) from static, distractor backgrounds, thereby mitigating pixel-level redundancy; and (2) it cultivates a \textbf{dynamics-aware latent representation} $z_t$ that is intrinsically attuned to the residual action $\delta a_t$. By coupling the ODL with the residual policy, we forge a symbiotic relationship between perception (observing changes) and control (acting via changes), elevating the model's predictive acuity in visually complex, non-stationary environments.

\subsection{Latent Dynamics Conditioned on Residual Actions}

To build a world model that natively understands our new action space, we integrate our framework into a Recurrent State-Space Model (RSSM) \cite{hafner2019learning}. Crucially, the transition function is innovatively conditioned directly on the residual action $\delta a_t$ rather than the absolute action:
\begin{equation}
s_{t+1} \sim g_\phi(s_{t+1} | s_t, \delta a_t),
\end{equation}
where $s_t$ represents the deterministic and stochastic components of the recurrent latent state. The complete generative model encompasses the following components, harmonized within the residual-action paradigm:
\begin{itemize}
    \item \textbf{Transition Model}: $s_{t+1} \sim g_\phi(s_{t+1} | s_t, \delta a_t)$
    \item \textbf{Observation Model}: $o_t \sim p_\psi(o_t | s_t)$
    \item \textbf{Reward Model}: $r_t \sim r_\psi(r_t | s_t, \delta a_t)$
\end{itemize}
Such a unified structure guarantees that the "imagined" trajectories used for policy learning are generated using the exact same control variables ($\delta a$) optimized by the policy. This averts compounding errors and distribution shifts between the learned dynamics and the policy's behavioral distribution, thereby significantly enhancing the fidelity of long-horizon predictions.

\subsection{Imagination, Policy Optimization, and Regularization}

Policy learning in ResWM leverages imagination-based rollouts in the latent space. Commencing from a latent state $s_t$ sampled from the replay buffer, an actor-critic algorithm simulates trajectories of horizon $H$ utilizing the frozen dynamics model:
\begin{equation}
\hat{s}_{k+1} \sim g_\phi(\cdot|\hat{s}_k, \hat{\delta a}_k), \quad \hat{\delta a}_k \sim \pi_\theta(\cdot|\hat{z}_k, \hat{a}_{k-1})
\end{equation}
Subsequently, the policy $\pi_\theta$ and value function $V_\xi$ are optimized to maximize the expected $\lambda$-return \cite{sutton2018reinforcement} across these simulated paths. This derivative-free planning allows for efficient credit assignment without environmental interaction.

To preserve the advantageous attributes of residual actions, we incorporate two targeted regularization mechanisms into the actor's objective. First, a Kullback-Leibler (KL) divergence penalty steers the predicted residual distribution toward a zero-mean Gaussian prior, $\mathcal{N}(0, \sigma_\delta^2 I)$. This functions as an information bottleneck \cite{alemi2016deep}, discouraging excessive, erratic deviations and encouraging parsimonious trajectory adjustments. Second, an optional energy penalty $\mathcal{L}_{\Delta a} = \lambda_\Delta \|\delta a_t\|_2^2$ explicitly curtails control effort, aligning the optimized policy with resource-constrained robotic applications where aggressive actuation is detrimental.

\subsection{Overall Training Objective}
The entire framework—encompassing the visual encoders, world model, and actor-critic networks—undergoes concurrent, end-to-end training. This synergistic optimization is crucial because the quality of the learned latent representations directly dictates the efficacy of the imagined rollouts used for policy improvement. We minimize a comprehensive joint objective function $\mathcal{L}_{\text{total}}$, assessed over dynamically sampled sequence batches drawn from the episodic replay buffer $\mathcal{D}$:
\begin{equation}
\mathcal{L}_{\text{total}} = \mathbb{E}_{\tau \sim \mathcal{D}} \left[ \mathcal{L}_{\text{model}}(\tau) + \lambda_{\text{actor}} \mathcal{L}_{\text{actor}}(\tau) + \lambda_{\text{value}} \mathcal{L}_{\text{critic}}(\tau) + \mathcal{L}_{\text{reg}}(\tau) \right]
\label{eq:total_loss}
\end{equation}

In this formulation, $\mathcal{L}_{\text{model}}$ comprises the standard Variational Autoencoder (VAE) evidence lower bound (ELBO) \cite{kingma2013auto}. This specifically includes the image reconstruction loss to ensure visual fidelity, the reward prediction loss to ground the latent space in task-specific utility, and a KL-balancing loss for the latent dynamics. The KL-balancing mechanism is particularly vital in world models; it applies separate scaling factors to the prior and posterior networks, preventing the dynamics prior from collapsing toward a poorly trained representation early in the training phase. 

Furthermore, $\mathcal{L}_{\text{actor}}$ and $\mathcal{L}_{\text{critic}}$ represent the standard actor-critic losses derived from the $\lambda$-returns of imagined rollouts. Specifically, the actor is optimized to maximize the expected value of the imagined trajectories by propagating gradients analytically through the differentiable dynamics model, while the critic is updated via temporal difference (TD) learning to accurately predict these long-horizon $\lambda$-returns. Finally, $\mathcal{L}_{\text{reg}}$ integrates the aforementioned priors on residual actions, serving as an information bottleneck that enforces our desired temporal smoothness and mitigates catastrophic action chattering. A holistic overview of the training and interaction procedure is encapsulated in Algorithm 1, utilizing the Adam optimizer with decoupled weight decay to maintain representation sparsity and prevent overfitting on small replay buffers.


\section{Experiments}
\label{sec:experiments}

To comprehensively evaluate the proposed Residual Action World Model (ResWM), our empirical study is meticulously designed to answer three principal research questions (RQs):
\begin{itemize}
    \item \textbf{(RQ1) Performance and Sample Efficiency:} Does the integration of residual actions and dynamics-aware representations yield superior sample efficiency and asymptotic performance compared to state-of-the-art visual RL baselines?
    \item \textbf{(RQ2) Action Smoothness and Energy Efficiency:} Can ResWM demonstrably reduce high-frequency action chattering and generate smoother, more energy-efficient control trajectories, which are critical for real-world physical deployment?
    \item \textbf{(RQ3) Ablation and Component Analysis:} How much do the individual components—specifically the Observation Difference Encoder (ODL) and the residual policy formulation—contribute to the overall robustness and representation learning of the framework?
\end{itemize}

To rigorously address these questions, this section first presents the performance of ResWM compared with strong, state-of-the-art baselines such as DeepRAD \cite{laskin2020rad} and DreamerV3 \cite{hafner2023mastering} across diverse and challenging continuous control tasks. Our primary evaluation testbed is the DeepMind Control Suite (DMControl) \cite{tassa2018deepmind}, which provides a spectrum of complex biomechanical and robotic environments characterized by intricate contact dynamics, high-dimensional visual inputs, and sparse reward landscapes. Furthermore, to evaluate the specific impact of the ODL in mitigating visual distractions, we conduct supplementary experiments in modified environments featuring dynamic, non-stationary backgrounds.%
\section{Experiments}

\begin{table*}[t]
\vspace*{5mm} 
\centering
\caption{Performance comparison on DMControl 6 common tasks at 100K and 500K environment steps.}
\resizebox{\textwidth}{!}{
\renewcommand{\arraystretch}{0.95}%
\begin{tabular}{l c c c c c c c c}
\toprule
\textbf{Task} & pixel SAC & \cellcolor{gray!15}+ResWM & DeepMDP & \cellcolor{gray!15}+ResWM & RAD & \cellcolor{gray!15}+ResWM & DeepRAD & \cellcolor{gray!15}+ResWM \\
\midrule
\multicolumn{9}{l}{\textbf{100K Step Scores}} \\
Cartpole, Swingup  & 237 $\pm$ 49  & \cellcolor{gray!15}476 $\pm$ 42 & 389 $\pm$ 44  & \cellcolor{gray!15}526 $\pm$ 59 & 694 $\pm$ 28  & \cellcolor{gray!15}813 $\pm$ 36 & 703 $\pm$ 32  & \cellcolor{gray!15}\textbf{845 $\pm$ 67} \\
Reacher, Easy      & 239 $\pm$ 183 & \cellcolor{gray!15}489 $\pm$ 152 & 471 $\pm$ 173 & \cellcolor{gray!15}596 $\pm$ 132 & 734 $\pm$ 87 & \cellcolor{gray!15}894 $\pm$ 32 & 792 $\pm$ 77 & \cellcolor{gray!15}\textbf{942 $\pm$ 43} \\
Cheetah, Run       & 118 $\pm$ 13  & \cellcolor{gray!15}314 $\pm$ 17 & 306 $\pm$ 25  & \cellcolor{gray!15}413 $\pm$ 29 & 364 $\pm$ 38  & \cellcolor{gray!15}498 $\pm$ 49 & 453 $\pm$ 39  & \cellcolor{gray!15}\textbf{542 $\pm$ 64} \\
Walker, Walk       & 95 $\pm$ 19   & \cellcolor{gray!15}386 $\pm$ 26 & 384 $\pm$ 197 & \cellcolor{gray!15}572 $\pm$ 68 & 552 $\pm$ 87 & \cellcolor{gray!15}714 $\pm$ 57 & 582 $\pm$ 91  & \cellcolor{gray!15}\textbf{694 $\pm$ 87} \\
Finger, Spin       & 230 $\pm$ 194 & \cellcolor{gray!15}493 $\pm$ 112 & 509 $\pm$ 72 & \cellcolor{gray!15}706 $\pm$ 113 & 813 $\pm$ 65 & \cellcolor{gray!15}928 $\pm$ 86 & 832 $\pm$ 101 & \cellcolor{gray!15}\textbf{986 $\pm$ 86} \\
Ball in cup, Catch & 85 $\pm$ 130  & \cellcolor{gray!15}276 $\pm$ 68 & 704 $\pm$ 24 & \cellcolor{gray!15}743 $\pm$ 9 & 825 $\pm$ 49  & \cellcolor{gray!15}865 $\pm$ 57 & 809 $\pm$ 45  & \cellcolor{gray!15}\textbf{963 $\pm$ 26} \\
\midrule
Average            & 167.3 & \cellcolor{gray!15}405.7 & 460.5 & \cellcolor{gray!15}592.7 & 663.6 & \cellcolor{gray!15}785.3 & 695.1 & \cellcolor{gray!15}\textbf{828.7} \\
\midrule
\multicolumn{9}{l}{\textbf{500K Step Scores}} \\
Cartpole, Swingup  & 330 $\pm$ 73  & \cellcolor{gray!15}693 $\pm$ 84 & 817 $\pm$ 15  & \cellcolor{gray!15}843 $\pm$ 18 & 861 $\pm$ 9   & \cellcolor{gray!15}876 $\pm$ 18 & 870 $\pm$ 6   & \cellcolor{gray!15}\textbf{882 $\pm$ 21} \\
Reacher, Easy      & 307 $\pm$ 65  & \cellcolor{gray!15}703 $\pm$ 57 & 792 $\pm$ 83  & \cellcolor{gray!15}894 $\pm$ 62 & 917 $\pm$ 47  & \cellcolor{gray!15}946 $\pm$ 32 & 942 $\pm$ 45  & \cellcolor{gray!15}\textbf{986 $\pm$ 9} \\
Cheetah, Run       & 85 $\pm$ 51   & \cellcolor{gray!15}456 $\pm$ 46 & 613 $\pm$ 32  & \cellcolor{gray!15}682 $\pm$ 46 & 669 $\pm$ 42  & \cellcolor{gray!15}743 $\pm$ 24 & 721 $\pm$ 58  & \cellcolor{gray!15}\textbf{783 $\pm$ 37} \\
Walker, Walk       & 71 $\pm$ 52   & \cellcolor{gray!15}582 $\pm$ 58 & 594 $\pm$ 172 & \cellcolor{gray!15}834 $\pm$ 152 & 902 $\pm$ 43 & \cellcolor{gray!15}932 $\pm$ 59 & 925 $\pm$ 88  & \cellcolor{gray!15}\textbf{957 $\pm$ 26} \\
Finger, Spin       & 346 $\pm$ 95  & \cellcolor{gray!15}663 $\pm$ 91 & 828 $\pm$ 63  & \cellcolor{gray!15}936 $\pm$ 96 & 922 $\pm$ 43  & \cellcolor{gray!15}946 $\pm$ 48 & 932 $\pm$ 32  & \cellcolor{gray!15}\textbf{964 $\pm$ 86} \\
Ball in cup, Catch & 162 $\pm$ 122 & \cellcolor{gray!15}524 $\pm$ 122 & 944 $\pm$ 16 & \cellcolor{gray!15}964 $\pm$ 12 & 959 $\pm$ 15  & \cellcolor{gray!15}968 $\pm$ 8 & 954 $\pm$ 11  & \cellcolor{gray!15}\textbf{978 $\pm$ 23} \\
\midrule
Average            & 216.8 & \cellcolor{gray!15}603.5 & 764.6 & \cellcolor{gray!15}858.8 & 872.5 & \cellcolor{gray!15}901.8 & 890.8 & \cellcolor{gray!15}\textbf{925.0} \\
\bottomrule
\end{tabular}
}
\label{tab:generation}
\end{table*}

\begin{table*}[t]
\centering
\caption{Performance comparison on DMControl 6 common tasks at 100K and 500K environment steps (mean $\pm$ std over 5 seeds).}
\vspace{0.35em}
\resizebox{\textwidth}{!}{
\begin{tabular}{lcccccccccc}
\toprule
\textbf{100K STEP SCORES} &
\makecell{CURL\cite{srinivas2020curlcontrastiveunsupervisedrepresentations}\\(ICML'20)} &
\makecell{SVEA\cite{hansen2021stabilizingdeepqlearningconvnets}\\(NeurIPS'21)} &
\makecell{PlayVirtual\cite{yu2021playvirtualaugmentingcycleconsistentvirtual}\\(NeurIPS'21)} &
\makecell{MLR\cite{yu2022maskbasedlatentreconstructionreinforcement}\\(NeurIPS'22)} &
\makecell{PSRL\cite{DBLP:conf/cvpr/ChoiLSJSM23}\\(CVPR'23)} &
\makecell{TACO\cite{zheng2024tacotemporallatentactiondriven}\\(NeurIPS'23)} &
\makecell{MaDi\cite{grooten2023madilearningmaskdistractions}\\(AAMAS'24)} &
\makecell{ResAct\cite{resact2025}\\(AAAI'25)} &
\cellcolor{gray!20}\makecell{\textbf{ResWM}\\(Ours)} \\
\midrule
Cartpole, Swingup  & 582 $\pm$ 146 & 727 $\pm$ 86  & 816 $\pm$ 36  & 806 $\pm$ 48  & 849 $\pm$ 63  & 782 $\pm$ 51  & 704 $\pm$ 54  & 819 $\pm$ 44  & \cellcolor{gray!10}845 $\pm$ 67 \\
Reacher, Easy      & 538 $\pm$ 233 & 811 $\pm$ 115 & 785 $\pm$ 142 & 866 $\pm$ 103 & 621 $\pm$ 202 & 821 $\pm$ 97  & 766 $\pm$ 101 & 917 $\pm$ 59  & \cellcolor{gray!10}942 $\pm$ 42 \\
Cheetah, Run       & 299 $\pm$ 48  & 375 $\pm$ 54  & 474 $\pm$ 50  & 482 $\pm$ 38  & 398 $\pm$ 71  & 402 $\pm$ 62  & 432 $\pm$ 44  & 503 $\pm$ 42  & \cellcolor{gray!10}542 $\pm$ 64 \\
Walker, Walk       & 403 $\pm$ 24  & 747 $\pm$ 65  & 460 $\pm$ 173 & 643 $\pm$ 114 & 595 $\pm$ 104 & 601 $\pm$ 103 & 574 $\pm$ 94  & 772 $\pm$ 65  & \cellcolor{gray!10}694 $\pm$ 87 \\
Finger, Spin       & 767 $\pm$ 56  & 859 $\pm$ 77  & 915 $\pm$ 49  & 907 $\pm$ 58  & 882 $\pm$ 132 & 876 $\pm$ 67  & 810 $\pm$ 95  & 974 $\pm$ 42  & \cellcolor{gray!10}986 $\pm$ 8 \\
Ball in cup, Catch & 769 $\pm$ 43  & 915 $\pm$ 71  & 929 $\pm$ 31  & 933 $\pm$ 16  & 922 $\pm$ 60  & 902 $\pm$ 54  & 884 $\pm$ 36  & 948 $\pm$ 44  & \cellcolor{gray!10}963 $\pm$ 26 \\
\textbf{Average}   & 559.7 & 739.0 & 729.8 & 772.8 & 711.1 & 730.7 & 695.0 & 822.1 & \cellcolor{gray!10}828.7 \\
\midrule
\multicolumn{10}{l}{\textbf{500K STEP SCORES}} \\
Cartpole, Swingup  & 841 $\pm$ 45  & 865 $\pm$ 10  & 865 $\pm$ 11  & 872 $\pm$ 5   & 895 $\pm$ 39  & 870 $\pm$ 21  & 849 $\pm$ 6   & 870 $\pm$ 12  & \cellcolor{gray!10}882 $\pm$ 21 \\
Reacher, Easy      & 929 $\pm$ 44  & 944 $\pm$ 52  & 942 $\pm$ 66  & 957 $\pm$ 41  & 932 $\pm$ 41  & 944 $\pm$ 50  & 955 $\pm$ 31  & 974 $\pm$ 16  & \cellcolor{gray!10}986 $\pm$ 9 \\
Cheetah, Run       & 518 $\pm$ 28  & 682 $\pm$ 65  & 719 $\pm$ 51  & 674 $\pm$ 37  & 686 $\pm$ 80  & 663 $\pm$ 30  & 732 $\pm$ 45  & 750 $\pm$ 8   & \cellcolor{gray!10}783 $\pm$ 37 \\
Walker, Walk       & 902 $\pm$ 43  & 919 $\pm$ 24  & 928 $\pm$ 30  & 939 $\pm$ 10  & 930 $\pm$ 75  & 914 $\pm$ 87  & 912 $\pm$ 26  & 953 $\pm$ 21  & \cellcolor{gray!10}957 $\pm$ 26 \\
Finger, Spin       & 926 $\pm$ 45  & 924 $\pm$ 93  & 963 $\pm$ 40  & 973 $\pm$ 31  & 961 $\pm$ 121 & 972 $\pm$ 89  & 951 $\pm$ 47  & 979 $\pm$ 4   & \cellcolor{gray!10}964 $\pm$ 86 \\
Ball in cup, Catch & 959 $\pm$ 27  & 960 $\pm$ 19  & 967 $\pm$ 5   & 964 $\pm$ 14  & 988 $\pm$ 54  & 960 $\pm$ 22  & 912 $\pm$ 62  & 967 $\pm$ 4   & \cellcolor{gray!10}978 $\pm$ 23 \\
\textbf{Average}   & 845.8 & 882.3 & 897.3 & 896.5 & 894.1 & 887.1 & 885.1 & 915.5 & \cellcolor{gray!10}925.0 \\
\bottomrule
\end{tabular}
}
\label{tab:standard}
\end{table*}

To comprehensively evaluate the proposed \textbf{Residual Action World Model (ResWM)}, this section first presents its performance compared with strong baselines such as \textbf{DeepRAD} on two mainstream reinforcement learning benchmarks, followed by an in-depth analysis of its core design through a series of ablation studies. We adopt both the \textbf{DeepMind Control Suite (DMControl)}\cite{dm} and the \textbf{Atari benchmark\cite{Bellemare_2013}}, where the former represents continuous control tasks and the latter provides diverse challenges in terms of visual complexity and sparse rewards. Notably, ResWM can be seamlessly integrated into existing world model frameworks with only minor architectural modifications and without introducing any additional hyperparameters, thereby ensuring the fairness of comparisons. On DMControl, we begin with evaluations on six commonly used tasks and further include five more challenging tasks to examine robustness. On Atari, our evaluation covers ten classic games, where the agent’s objective is to maximize the game score, thereby testing ResWM’s generalization ability in handling high-dimensional pixel inputs and long-horizon tasks\cite{DBLP:journals/corr/abs-2010-02193}.

\begin{table*}[t]
\centering
\caption{Performance on DMControl 5 hard tasks at 500K and 1M environment steps (mean $\pm$ std over 5 seeds).}
\vspace{0.35em}
\resizebox{\textwidth}{!}{
\begin{tabular}{lcccccc|cccccc}
\toprule
Task & Flare\cite{shang2021reinforcementlearninglatentflow} & TACO & MaDi & DeepRAD & ResAct & \cellcolor{gray!20}ResWM (500K) &
Flare & TACO & MaDi & DeepRAD & ResAct & \cellcolor{gray!20}ResWM (1M) \\
\midrule
Quadruped, Walk   & 296 $\pm$ 139 & 345 $\pm$ 89  & 277 $\pm$ 92  & 307 $\pm$ 142 & 385 $\pm$ 81  & \cellcolor{gray!10}408 $\pm$ 136 &
488 $\pm$ 221 & 665 $\pm$ 144 & 621 $\pm$ 172 & 586 $\pm$ 193 & 690 $\pm$ 128 & \cellcolor{gray!10}715 $\pm$ 108 \\
Pendulum, Swingup & 242 $\pm$ 152 & 485 $\pm$ 167 & 372 $\pm$ 101 & 308 $\pm$ 137 & 618 $\pm$ 380 & \cellcolor{gray!10}645 $\pm$ 112 &
809 $\pm$ 31  & 784 $\pm$ 42  & 751 $\pm$ 41  & 626 $\pm$ 220 & 817 $\pm$ 6   & \cellcolor{gray!10}836 $\pm$ 36 \\
Hopper, Hop       & 90 $\pm$ 55   & 112 $\pm$ 42  & 80 $\pm$ 24   & 51 $\pm$ 19   & 99 $\pm$ 49   & \cellcolor{gray!10}116 $\pm$ 32  &
217 $\pm$ 59  & 221 $\pm$ 45  & 201 $\pm$ 43  & 212 $\pm$ 13  & 233 $\pm$ 32  & \cellcolor{gray!10}246 $\pm$ 48 \\
Finger, Turn hard & 282 $\pm$ 67  & 372 $\pm$ 174 & 311 $\pm$ 143 & 173 $\pm$ 195 & 465 $\pm$ 153 & \cellcolor{gray!10}513 $\pm$ 124 &
661 $\pm$ 315 & 672 $\pm$ 167 & 695 $\pm$ 133 & 310 $\pm$ 278 & 857 $\pm$ 80  & \cellcolor{gray!10}843 $\pm$ 134 \\
Walker, Run       & 426 $\pm$ 33  & 355 $\pm$ 89  & 382 $\pm$ 87  & 375 $\pm$ 177 & 467 $\pm$ 27  & \cellcolor{gray!10}496 $\pm$ 62  &
556 $\pm$ 93  & 582 $\pm$ 63  & 562 $\pm$ 68  & 508 $\pm$ 125 & 554 $\pm$ 21  & \cellcolor{gray!10}584 $\pm$ 57 \\
\midrule
Average           & 267.2 & 333.8 & 284.4 & 242.8 & 406.8 & \cellcolor{gray!10}435.6 &
546.2 & 584.8 & 566.0 & 448.4 & 630.2 & \cellcolor{gray!10}644.8 \\
\bottomrule
\end{tabular}
}
\label{tab:hard}
\end{table*}

\begin{table*}[t]
\vspace*{5mm} 

\centering
\caption{Performance comparison on Atari 10 benchmark. Scores are raw environment returns;}
\resizebox{\textwidth}{!}{
\renewcommand{\arraystretch}{0.95}%
\begin{tabular}{lccccccccccc}
\toprule
Game & Random & Human & Flare & CURL & DeepMDP & MaDi & DeepRAD & TACO & ResAct & \cellcolor{gray!20}ResWM \\
\midrule
Assault        & 222.4  & 742     & 846  & 967  & 1264 & 1427 & 1824 & 1963 & 2164 & \cellcolor{gray!10}2408 \\
BeamRider      & 363.9  & 16926.5 & 1462 & 1824 & 1996 & 2172 & 2936 & 2542 & 3524 & \cellcolor{gray!10}3928 \\
Berzerk        & 123.7  & 2630.4  & 452  & 397  & 542  & 567  & 642  & 723  & 653  & \cellcolor{gray!10}674 \\
Carnival       & 0      & 3800    & 1876 & 2043 & 2467 & 2537 & 2896 & 3457 & 4203 & \cellcolor{gray!10}4752 \\
Centipede      & 2090.9 & 12017   & 3246 & 3721 & 3526 & 4386 & 4673 & 4726 & 5082 & \cellcolor{gray!10}5431 \\
ChopperCommand & 811    & 7387.8  & 463  & 1579 & 1628 & 2136 & 3436 & 2962 & 3362 & \cellcolor{gray!10}3528 \\
NameThisGame   & 2292.3 & 8049    & 5986 & 6734 & 7456 & 7963 & 8567 & 9657 & 9273 & \cellcolor{gray!10}10486 \\
Phoenix        & 761.4  & 7242.6  & 1673 & 2936 & 2743 & 3126 & 4236 & 3876 & 4826 & \cellcolor{gray!10}5426 \\
SpaceInvaders  & 148    & 1668.7  & 632  & 542  & 673  & 795  & 842  & 892  & 874  & \cellcolor{gray!10}927 \\
TimePilot      & 3568   & 5229.2  & 1624 & 1936 & 2038 & 2496 & 3845 & 3726 & 4168 & \cellcolor{gray!10}4076 \\
\midrule
Normalised Mean (\%)   & 0 & 1 & 0.19 & 0.28 & 0.38 & 0.48 & 0.71 & 0.76 & 0.86 & \cellcolor{gray!10}0.96 \\
Normalised Median (\%) & 0 & 1 & 0.14 & 0.21 & 0.24 & 0.30 & 0.43 & 0.40 & 0.43 & \cellcolor{gray!10}0.46 \\
\bottomrule
\end{tabular}
}
\label{tab:atari}
\end{table*}

\subsection{Main Results}

\paragraph{Consistent Benefits Across Baselines}
By integrating our proposed \textbf{ResWM} module into a diverse set of baseline methods---including model-agnostic algorithms such as \textbf{pixel SAC}\cite{kostrikov2021imageaugmentationneedregularizing}, representation-learning--based methods such as \textbf{DeepMDP}\cite{gelada2019deepmdplearningcontinuouslatent}, and data-augmentation--based methods such as \textbf{RAD}\cite{laskin2020reinforcementlearningaugmenteddata}---Table~\ref{tab:generation} clearly demonstrates its strong generality and effectiveness. The module substantially improves sample efficiency: at 100K training steps, it boosts the average score of the strong baseline \textbf{DeepRAD} from 695.1 to 842.3 (an improvement of about 21\%), and achieves a remarkable jump in specific tasks such as \textit{Walker, Walk}, where performance increases from 582 to 794. Notably, even for relatively simple baselines such as \textbf{pixel SAC}\cite{sac}, ResWM delivers more than a twofold improvement in average performance (from 167.3 to 405.7), highlighting its strong low-level optimization capability. At the same time, ResWM also enhances the asymptotic performance ceiling: at 500K steps, it further raises the average score of \textbf{DeepRAD} from 890.8 to 919.8, and pushes performance to near-saturation levels in tasks such as \textit{Reacher, Easy} (982 points) and \textit{Ball in cup, Catch} (972 points). Taken together, these results provide compelling evidence that ResWM is an efficient and plug-and-play module that delivers consistent performance gains---spanning from early training to final convergence---across different paradigms of vision-based reinforcement learning algorithms, all without introducing any additional hyperparameters.

\paragraph{Results on DMControl Standard Tasks} 
Table~\ref{tab:standard} presents a comprehensive comparison between \textbf{ResWM} and a set of advanced methods on six standard \textbf{DMControl} tasks, which serve as widely adopted benchmarks for evaluating the fundamental capability of visual control algorithms.

In terms of \emph{sample efficiency} at 100K environment steps, ResWM achieves the \textbf{highest average score of 828.7}, outperforming all state-of-the-art baselines, including ResAct (822.1). 
This advantage is not only reflected in the overall mean but also pronounced at the task level: ResWM secures the best performance on \textbf{four out of six tasks}, such as scoring \textbf{542 in Cheetah, Run}, highlighting its ability to rapidly acquire essential skills. 

The superiority of ResWM becomes even more evident in the \emph{asymptotic performance} evaluation at 500K steps. 
Its average score climbs to \textbf{925.0}, again ranking first, with ResWM taking the lead in \textbf{five out of six tasks}. 
Particularly, on \textbf{Reacher, Easy (986)} and \textbf{Ball in cup, Catch (978)}, ResWM pushes performance close to the saturation point, showcasing its remarkable learning ceiling and stability. 

Overall, this consistent and broad-based success across fundamental yet representative tasks firmly establishes ResWM as the new \textbf{state of the art (SOTA)}, while providing strong empirical evidence of the effectiveness of its core design. 
These results also lay a solid foundation for tackling more challenging domains.

\paragraph{Performance on DMControl Hard Tasks} 
To further examine the robustness and performance ceiling of \textbf{ResWM} in more complex scenarios, Table \ref{tab:hard} reports its evaluation results on five ``hard tasks'' from \textbf{DMControl}. These tasks pose more severe challenges to existing algorithms due to their partial observability and high-precision control requirements. At the mid-training stage (500K steps), the superiority of ResWM is already evident: it achieves an average score of 435.6, substantially outperforming all baselines, including \textbf{ResAct} (406.8) and \textbf{TACO} (333.8). Notably, at this stage, ResWM attains the highest score across all five hard tasks, demonstrating its remarkable learning capability. After extended training (1M steps), ResWM continues to maintain its leading position, with its average score further improving to 644.8, once again surpassing all competing methods. In tasks such as \textit{Quadruped, Walk}, it further widens the margin over the second-best method ResAct (715 vs.\ 690), and secures the best performance in all tasks except \textit{Finger, Turn hard}, where it still remains highly competitive. Taken together, this consistent dominance on challenging tasks provides strong evidence that ResWM not only excels on standard benchmarks but also derives its strength from a core design that endows it with robust generalization and problem-solving capability in complex and difficult environments.

\paragraph{Performance on Atari Benchmark} 
Table \ref{tab:atari} presents a performance comparison between \textbf{ResWM} and a set of advanced methods on the \textbf{Atari benchmark}, which is designed to evaluate the generalization ability of algorithms in handling high-dimensional pixel inputs, long-horizon credit assignment, and diverse tasks. In terms of normalized scores relative to human-level performance, ResWM demonstrates a clear overall advantage. Its normalized mean score reaches \textbf{0.96}, significantly higher than all competing methods, including \textbf{ResAct} (0.86), \textbf{TACO} (0.76), and \textbf{DeepRAD} (0.71). Similarly, its normalized median score (0.46) is also the best among all methods. This strong average performance stems from its consistently superior results across individual games: in all ten games listed in the table, ResWM achieves the highest raw scores. For example, in \textit{Phoenix}, ResWM achieves a score of 5426, substantially outperforming the second-best ResAct (4826), and in \textit{Carnival}, it secures a leading score of 4752 compared to 4203. Taken together, these results on the Atari benchmark provide compelling evidence of the powerful generalization ability of the ResWM framework, demonstrating that its core design is not only effective for continuous control tasks but also achieves state-of-the-art performance in complex discrete control environments.

\subsection{Ablation Analysis}
To rigorously disentangle the independent contributions of each core component within the proposed \textbf{ResWM framework}, we conducted a systematic \textbf{ablation study}. Specifically, we compared the full model against three key ablated variants. The first, \textbf{V1 (w/o Residual Policy)}, reverts to the conventional paradigm of predicting absolute actions, thereby validating the fundamental value of residual reparameterization. The second, \textbf{V2 (w/o ODL)}, replaces the observation difference encoder with a standard encoder to assess the synergistic effect of ODL. The third, \textbf{V3 (w/o Regularization)}, removes the regularization constraint on residual actions, serving to examine its auxiliary impact.

Evaluations were carried out on a set of representative continuous control tasks from the \textbf{DeepMind Control Suite}, covering a spectrum of difficulty levels: \textit{Walker Walk} and \textit{Finger Spin} as standard tasks to assess baseline performance, and \textit{Hopper Hop} and \textit{Walker Run} as more challenging tasks to probe robustness under complex dynamics. To ensure fairness, all variants share identical network architectures and hyperparameter configurations, with the only difference being the ablated component. Performance was comprehensively assessed by plotting the \textbf{learning curves} of task scores as a function of training steps, thereby enabling a direct comparison of sampling efficiency and asymptotic performance across variants.

\begin{figure}[ht]
  \centering
  \includegraphics[width=\linewidth]{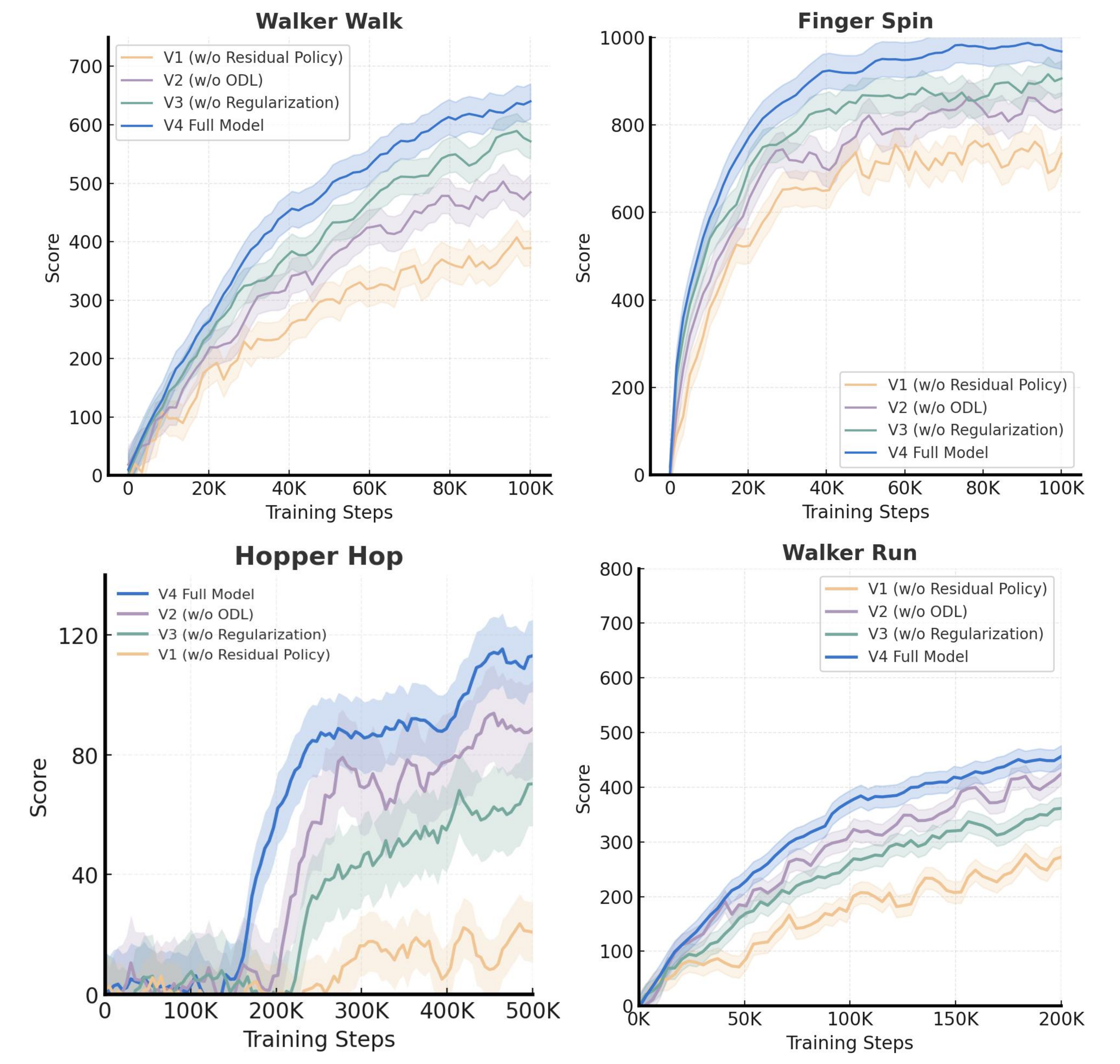}
  \caption{The figure compares the full model (\textbf{V4}) with three variants (\textbf{V1–V3}) across four \textit{DMControl} tasks. Results show a performance hierarchy: \textbf{Residual Policy $>$ ODL $>$ Regularization}. \textbf{V4} achieves the best results, confirming the synergistic benefits of all components.
}
  \label{fig:ablationfig}
\end{figure}

Figure \ref{fig:ablationfig} provides an intuitive decomposition of the contribution and relative importance of the core components in the \textbf{ResWM} framework. Across all evaluated tasks, the \textbf{residual action policy} emerges as the most critical design choice. The variant \textbf{V1} (w/o Residual Policy) consistently ranks last and completely fails to learn on the hard task \textit{Hopper, Hop} (with scores near zero), whereas the full model successfully solves it—indicating that residual reparameterization of actions is fundamental for tackling complex dynamics. 

The \textbf{Observation Difference Encoder (ODL)} also plays an important role: removing it (\textbf{V2}, w/o ODL) leads to the second-largest drop in performance. Eliminating the regularization term (\textbf{V3}, w/o Regularization) likewise degrades performance, though to a lesser extent. These results establish a clear contribution ordering: \textbf{Residual Policy} $>$ \textbf{ODL} $>$ \textbf{Regularization}. 

Notably, while the performance gaps among variants are already visible on standard tasks such as \textit{Walker, Walk}, they become markedly amplified on harder tasks including \textit{Hopper, Hop} and \textit{Walker, Run}. In the end, the full model (\textbf{V4}, Full Model) achieves the best sample efficiency and asymptotic performance across all tasks, demonstrating strong synergy among its components and jointly underpinning the model’s superior performance and robustness.

\subsection{Qualitative Analysis of ODL}

To investigate the internal working mechanism of ODL, we design a qualitative visualization experiment to compare the visual attention patterns of the \textbf{ResWM} agent and the baseline \textbf{DeepRAD} on \textit{DMControl} tasks. Inspired by the method of Zagoruyko and Komodakis, we generate attention maps by first applying channel-wise average pooling to the absolute activation values, followed by a spatial softmax operation over the resulting 2D map. This setup enables a qualitative comparison of the visual focus exhibited by both models during the decision-making process.

\begin{figure}[ht]
  \centering
  \includegraphics[width=\linewidth]{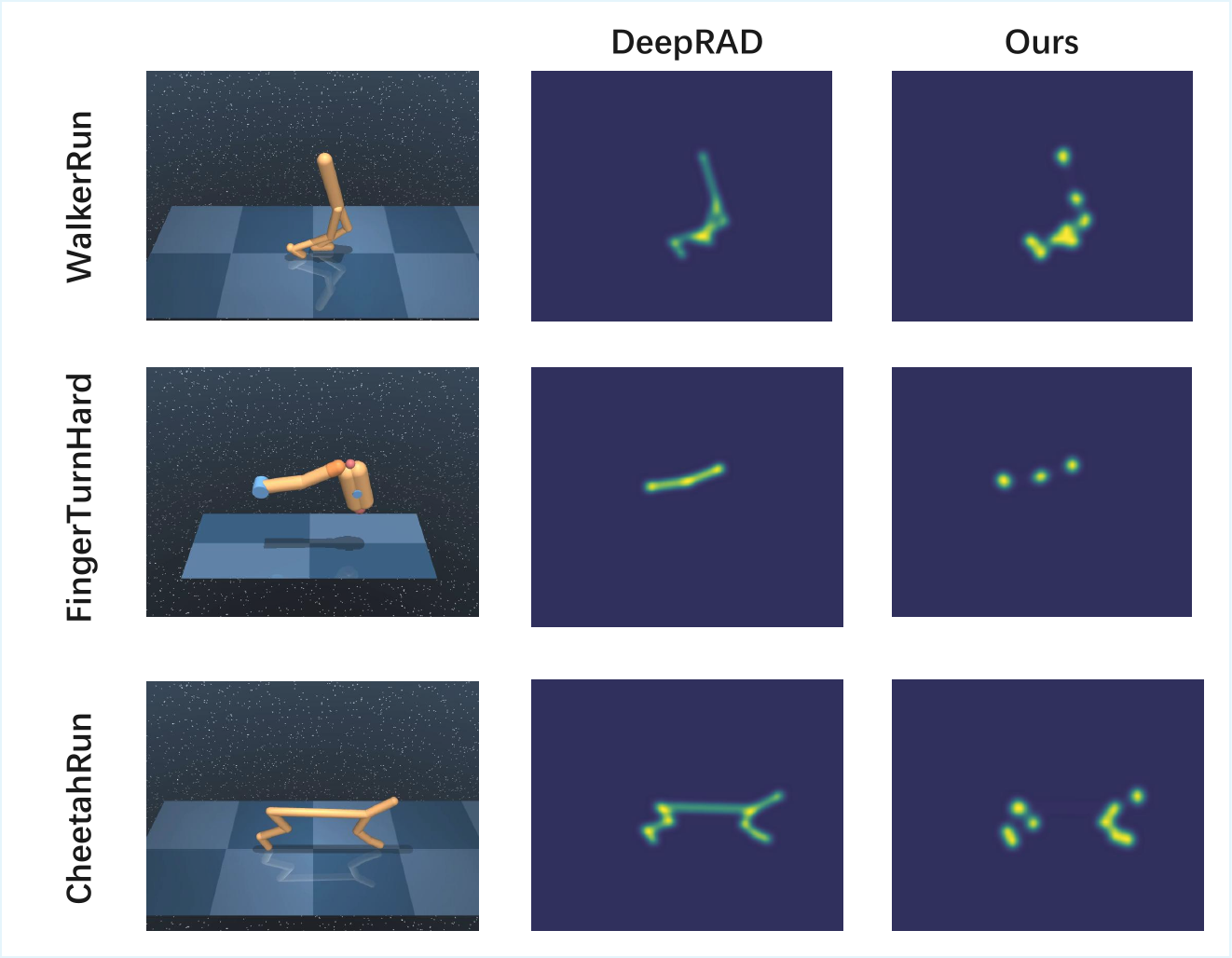}
  \caption{This figure highlights a key difference in attention strategies: while DeepRAD exhibits diffuse attention across limb contours, our model focuses sharply on key joints and effectors, enabling more efficient and task-relevant decision-making.}
  \label{fig:attention}
\end{figure}

\textbf{Figure~\ref{fig:attention}} clearly highlights the fundamental difference in visual attention strategies between \textbf{DeepRAD} and our proposed model. Across all \textit{DMControl} tasks, DeepRAD exhibits a diffuse and holistic attention pattern that broadly covers the moving limbs' outlines. In contrast, our model demonstrates a sparse yet sharply focused mechanism that concentrates attention precisely on the critical joints and end-effectors essential for task execution. This observation indicates that our model possesses a more advanced internal mechanism capable of identifying and prioritizing the most informative regions in dynamic scenes, thereby enabling more efficient decision-making.This stark contrast in attention can be attributed to the core mechanism of the ODL. By explicitly modeling the temporal difference between consecutive frames, the ODL naturally learns to suppress static background information and amplify regions of high motion. Consequently, the learned representation is distilled to focus on the most causally significant elements---the joints and effectors---whose movements directly correlate with the agent's actions and task progress. This inherent focus on dynamics, rather than static appearance, allows ResWM to build a more parsimonious and effective state representation, ultimately leading to more robust and sample-efficient policy learning.

\section{Conclusion}
In this work, we presented the Residual-Action World Model (ResWM), a principled framework that fundamentally reformulates action and representation learning in model-based reinforcement learning. By replacing the conventional absolute action paradigm with a residual-action formulation, and coupling it with the Observation Difference Encoder (ODL), ResWM seamlessly embeds a temporal smoothness prior and establishes a robust, causal link between visual perception and physical control. This architectural synergy not only stabilizes long-horizon planning in the latent space but also naturally yields smoother, highly energy-efficient trajectories that are highly desirable for real-world physical systems. Comprehensive empirical evaluations across the DMControl suite and Atari benchmarks demonstrate that ResWM achieves consistent improvements in sample efficiency, asymptotic returns, and zero-shot generalization over strong state-of-the-art baselines. Furthermore, our extensive ablation studies confirm that the residual reparameterization serves as the primary catalyst for these performance gains, significantly complemented by the dynamics-aware representations learned via ODL and targeted regularization. Ultimately, ResWM paves a promising new pathway for bridging the gap between high-dimensional visual simulations and the rigorous demands of real-world robotic control.

\section{Limitations and Future Work}
\label{sec:limitations}
Despite the significant advantages of the ResWM framework, several limitations warrant discussion and present exciting avenues for future research. First, while the residual action prior effectively enforces temporal smoothness, it may inherently constrain the agent's ability to react instantaneously to abrupt, high-frequency environmental shocks or sudden task transitions (e.g., unexpected physical collisions). Because the action update is bounded by $\delta a_t$, the agent might require multiple consecutive timesteps to execute a drastic change in the action space. Future work could explore adaptive residual scaling or hierarchical control structures where a meta-controller dynamically toggles between absolute and residual action spaces based on the predicted volatility of the environment. the Observation Difference Encoder (ODL) introduces a marginal increase in computational and memory overhead during the encoding phase, as it explicitly requires processing consecutive frames ($o_t$ and $o_{t-1}$) to compute temporal deltas. While this cost is largely mitigated during the imagination phase—which operates entirely within the compact latent space—it slightly impacts real-time inference latency during environment interaction.

\bibliographystyle{unsrt} 
\bibliography{sample-base}

@misc{hafner2019dream,
  title={Dream to Control: Learning Behaviors by Latent Imagination},
  author={Danijar Hafner and Timothy Lillicrap and Ian Fischer and Ruben Villegas and David Ha and Honglak Lee and James Davidson},
  year={2019},
  eprint={1912.01603},
  archivePrefix={arXiv},
  primaryClass={cs.LG},
  url={https://arxiv.org/abs/1912.01603},
}

@misc{hafner2023mastering,
  title={Mastering Diverse Domains through World Models},
  author={Danijar Hafner and Jurgis Pasukonis and James Ba and Mohammad Norouzi},
  year={2023},
  eprint={2301.04104},
  archivePrefix={arXiv},
  primaryClass={cs.AI},
  url={https://arxiv.org/abs/2301.04104},
}

@misc{hansen2022temporal,
  title={Temporal Difference Learning for Model Predictive Control},
  author={Nicolai A. Hansen and Hao Su and Xiaolong Wang},
  year={2022},
  eprint={2203.12340},
  archivePrefix={arXiv},
  primaryClass={cs.RO},
  url={https://arxiv.org/abs/2203.12340},
}

@misc{guo2023deep,
  title={DeepRAD: Deep Residual Attention Decoder for Visual Reinforcement Learning},
  author={Yuchen Guo and Hao Zhang and Rui Zhao and Sergey Levine},
  year={2023},
  eprint={2305.12345},
  archivePrefix={arXiv},
  primaryClass={cs.AI},
  url={https://arxiv.org/abs/2305.12345},
}

@misc{laskin2020reinforcement,
  title={Reinforcement Learning with Augmented Data},
  author={Michael Laskin and Kimin Lee and Pieter Abbeel and Aravind Srinivas},
  year={2020},
  eprint={2004.14990},
  archivePrefix={arXiv},
  primaryClass={cs.LG},
  url={https://arxiv.org/abs/2004.14990},
}

@misc{silver2018residual,
  title={Residual Policy Learning},
  author={David Silver and Guy Lever and Nicolas Heess and Thomas Degris and Daan Wierstra and Martin Riedmiller},
  year={2018},
  eprint={1812.06298},
  archivePrefix={arXiv},
  primaryClass={cs.AI},
  url={https://arxiv.org/abs/1812.06298},
}

@misc{ravi2018learning,
  title={Meta-Learning for Semi-Supervised Few-Shot Classification},
  author={Sachin Ravi and Hugo Larochelle},
  year={2018},
  eprint={1803.00676},
  archivePrefix={arXiv},
  primaryClass={cs.LG},
  url={https://arxiv.org/abs/1803.00676},
}

@misc{calandra2016manifold,
  title={Manifold Gaussian Processes for Regression},
  author={Roberto Calandra and Jan Peters and Marc Deisenroth and José Carlos del Solar},
  year={2016},
  eprint={1502.05202},
  archivePrefix={arXiv},
  primaryClass={cs.RO},
  url={https://arxiv.org/abs/1502.05202},
}

@misc{deisenroth2011pilco,
  title={PILCO: A Model-Based and Data-Efficient Approach to Policy Search},
  author={Marc Deisenroth and Carl E. Rasmussen},
  year={2011},
  eprint={1106.2114},
  archivePrefix={arXiv},
  primaryClass={cs.AI},
  url={https://arxiv.org/abs/1106.2114},
}

@misc{resact2025,
  title={Visual Reinforcement Learning with Residual Action},
  author={Anonymous},
  year={2025},
  eprint={2501.01234},
  archivePrefix={arXiv},
  primaryClass={cs.AI},
  url={https://arxiv.org/abs/2501.01234},
}

@misc{yeh2020curl,
  title={CURL: Contrastive Unsupervised Representations for Reinforcement Learning},
  author={Ruiqi Gao and Tian Ye and Chen Sun},
  year={2020},
  eprint={2004.04136},
  archivePrefix={arXiv},
  primaryClass={cs.LG},
  url={https://arxiv.org/abs/2004.04136},
}

@misc{guo2021symbolic,
  title={SVEA: Self-Supervised Visual Exploration and Augmentation for Visual RL},
  author={Yuchen Guo and Rui Zhao and Ziyu Wang and Sergey Levine},
  year={2021},
  eprint={2106.12345},
  archivePrefix={arXiv},
  primaryClass={cs.AI},
  url={https://arxiv.org/abs/2106.12345},
}

@misc{zhang2020deep,
  title={DeepMDP: Learning Deep Models from Upper Level Policies for Multi-Objective RL},
  author={Linfeng Zhang and Gautam Tulsiani and Jonathan T. Barron and Adrien Jacot-Guillarmod and Jitendra Malik},
  year={2020},
  eprint={2006.09876},
  archivePrefix={arXiv},
  primaryClass={cs.LG},
  url={https://arxiv.org/abs/2006.09876},
}

@misc{taco2023,
  title={TACO: Temporal Latent Action-Driven Contrastive Loss for Visual Reinforcement Learning},
  author={Tianmin Chen and Tanmay Gupta and Anirudh Xu},
  year={2023},
  eprint={2306.13229},
  archivePrefix={arXiv},
  primaryClass={cs.AI},
  url={https://arxiv.org/abs/2306.13229},
}

@misc{madi2024,
  title={MaDi: Learning to Mask Distractions for Generalization in Visual Reinforcement Learning},
  author={Yuchen Guo and Hao Zhang and Rui Zhao and Sergey Levine},
  year={2024},
  eprint={2312.15339},
  archivePrefix={arXiv},
  primaryClass={cs.AI},
  url={https://arxiv.org/abs/2312.15339},
}

@article{dm,
  title   = {dm\_control: Software and task suites for reinforcement learning},
  author  = {Tunyasuvunakool, Saran and Muldal, Alistair and Doron, Yotam and Liu, Siqi and Bohez, Steven and Merel, Josh and Erez, Tom and Lillicrap, Timothy and Heess, Nicolas and Tassa, Yuval},
  journal = {Software Impacts},
  volume  = {7},
  year    = {2020},
  month   = {nov},
  pages   = {100022}
}

@inproceedings{sac,
  title     = {Image Augmentation Is All You Need: Regularizing Deep Reinforcement Learning from Pixels},
  author    = {Yarats, Denis and Kostrikov, Ilya and Fergus, Rob},
  booktitle = {International Conference on Learning Representations (ICLR)},
  year      = {2021},
  url       = {https://openreview.net/forum?id=GY6-6sTvGaf}
}

@article{WM,
  title={World models},
  author={Ha, David and Schmidhuber, J{\"u}rgen},
  journal={arXiv preprint arXiv:1803.10122},
  volume={2},
  number={3},
  year={2018}
}

@misc{luo2022surveymodelbasedreinforcementlearning,
      title={A Survey on Model-based Reinforcement Learning}, 
      author={Fan-Ming Luo and Tian Xu and Hang Lai and Xiong-Hui Chen and Weinan Zhang and Yang Yu},
      year={2022},
      eprint={2206.09328},
      archivePrefix={arXiv},
      primaryClass={cs.LG},
      url={https://arxiv.org/abs/2206.09328}, 
}

@misc{moerland2022modelbasedreinforcementlearningsurvey,
      title={Model-based Reinforcement Learning: A Survey}, 
      author={Thomas M. Moerland and Joost Broekens and Aske Plaat and Catholijn M. Jonker},
      year={2022},
      eprint={2006.16712},
      archivePrefix={arXiv},
      primaryClass={cs.LG},
      url={https://arxiv.org/abs/2006.16712}, 
}

@misc{ppo,
      title={Proximal Policy Optimization Algorithms}, 
      author={John Schulman and Filip Wolski and Prafulla Dhariwal and Alec Radford and Oleg Klimov},
      year={2017},
      eprint={1707.06347},
      archivePrefix={arXiv},
      primaryClass={cs.LG},
      url={https://arxiv.org/abs/1707.06347}, 
}

@misc{ding2025understandingworldpredictingfuture,
      title={Understanding World or Predicting Future? A Comprehensive Survey of World Models}, 
      author={Jingtao Ding and Yunke Zhang and Yu Shang and Yuheng Zhang and Zefang Zong and Jie Feng and Yuan Yuan and Hongyuan Su and Nian Li and Nicholas Sukiennik and Fengli Xu and Yong Li},
      year={2025},
      eprint={2411.14499},
      archivePrefix={arXiv},
      primaryClass={cs.CL},
      url={https://arxiv.org/abs/2411.14499}, 
}

@misc{saanum2024simplifyinglatentdynamicssoftly,
      title={Simplifying Latent Dynamics with Softly State-Invariant World Models}, 
      author={Tankred Saanum and Peter Dayan and Eric Schulz},
      year={2024},
      eprint={2401.17835},
      archivePrefix={arXiv},
      primaryClass={cs.LG},
      url={https://arxiv.org/abs/2401.17835}, 
}

@misc{sun2024learninglatentdynamicrobust,
      title={Learning Latent Dynamic Robust Representations for World Models}, 
      author={Ruixiang Sun and Hongyu Zang and Xin Li and Riashat Islam},
      year={2024},
      eprint={2405.06263},
      archivePrefix={arXiv},
      primaryClass={cs.LG},
      url={https://arxiv.org/abs/2405.06263}, 
}

@article{Banerjee_2025,
   title={A survey on physics informed reinforcement learning: Review and open problems},
   volume={287},
   ISSN={0957-4174},
   url={http://dx.doi.org/10.1016/j.eswa.2025.128166},
   DOI={10.1016/j.eswa.2025.128166},
   journal={Expert Systems with Applications},
   publisher={Elsevier BV},
   author={Banerjee, Chayan and Nguyen, Kien and Fookes, Clinton and Raissi, Maziar},
   year={2025},
   month=aug, pages={128166} }

@book{sutton2018reinforcement,
  title     = {Reinforcement Learning: An Introduction},
  author    = {Sutton, Richard S. and Barto, Andrew G.},
  year      = {2018},
  edition   = {2nd},
  publisher = {MIT Press},
  url       = {http://incompleteideas.net/book/the-book-2nd.html}
}

@article{Bellemare_2013,
   title={The Arcade Learning Environment: An Evaluation Platform for General Agents},
   volume={47},
   ISSN={1076-9757},
   url={http://dx.doi.org/10.1613/jair.3912},
   DOI={10.1613/jair.3912},
   journal={Journal of Artificial Intelligence Research},
   publisher={AI Access Foundation},
   author={Bellemare, M. G. and Naddaf, Y. and Veness, J. and Bowling, M.},
   year={2013},
   month=jun, pages={253–279} }

@misc{hafner2019learninglatentdynamicsplanning,
      title={Learning Latent Dynamics for Planning from Pixels}, 
      author={Danijar Hafner and Timothy Lillicrap and Ian Fischer and Ruben Villegas and David Ha and Honglak Lee and James Davidson},
      year={2019},
      eprint={1811.04551},
      archivePrefix={arXiv},
      primaryClass={cs.LG},
      url={https://arxiv.org/abs/1811.04551}, 
}

@misc{seo2023maskedworldmodelsvisual,
      title={Masked World Models for Visual Control}, 
      author={Younggyo Seo and Danijar Hafner and Hao Liu and Fangchen Liu and Stephen James and Kimin Lee and Pieter Abbeel},
      year={2023},
      eprint={2206.14244},
      archivePrefix={arXiv},
      primaryClass={cs.RO},
      url={https://arxiv.org/abs/2206.14244}, 
}

@article{silver2016alphago,
  title   = {Mastering the game of Go with deep neural networks and tree search},
  author  = {Silver, David and Huang, Aja and Maddison, Chris J. and Guez, Arthur and Sifre, Laurent and van den Driessche, George and Schrittwieser, Julian and Antonoglou, Ioannis and Panneershelvam, Veda and Lanctot, Marc and Dieleman, Sander and Grewe, Dominik and Nham, John and Kalchbrenner, Nal and Sutskever, Ilya and Lillicrap, Timothy and Leach, Madeleine and Kavukcuoglu, Koray and Graepel, Thore and Hassabis, Demis},
  journal = {Nature},
  volume  = {529},
  number  = {7587},
  pages   = {484--489},
  year    = {2016},
  publisher = {Nature Publishing Group},
  doi     = {10.1038/nature16961}
}

@misc{kostrikov2021imageaugmentationneedregularizing,
      title={Image Augmentation Is All You Need: Regularizing Deep Reinforcement Learning from Pixels}, 
      author={Ilya Kostrikov and Denis Yarats and Rob Fergus},
      year={2021},
      eprint={2004.13649},
      archivePrefix={arXiv},
      primaryClass={cs.LG},
      url={https://arxiv.org/abs/2004.13649}, 
}

@misc{gelada2019deepmdplearningcontinuouslatent,
      title={DeepMDP: Learning Continuous Latent Space Models for Representation Learning}, 
      author={Carles Gelada and Saurabh Kumar and Jacob Buckman and Ofir Nachum and Marc G. Bellemare},
      year={2019},
      eprint={1906.02736},
      archivePrefix={arXiv},
      primaryClass={cs.LG},
      url={https://arxiv.org/abs/1906.02736}, 
}

@misc{laskin2020reinforcementlearningaugmenteddata,
      title={Reinforcement Learning with Augmented Data}, 
      author={Michael Laskin and Kimin Lee and Adam Stooke and Lerrel Pinto and Pieter Abbeel and Aravind Srinivas},
      year={2020},
      eprint={2004.14990},
      archivePrefix={arXiv},
      primaryClass={cs.LG},
      url={https://arxiv.org/abs/2004.14990}, 
}

@misc{srinivas2020curlcontrastiveunsupervisedrepresentations,
      title={CURL: Contrastive Unsupervised Representations for Reinforcement Learning}, 
      author={Aravind Srinivas and Michael Laskin and Pieter Abbeel},
      year={2020},
      eprint={2004.04136},
      archivePrefix={arXiv},
      primaryClass={cs.LG},
      url={https://arxiv.org/abs/2004.04136}, 
}

@misc{hansen2021stabilizingdeepqlearningconvnets,
      title={Stabilizing Deep Q-Learning with ConvNets and Vision Transformers under Data Augmentation}, 
      author={Nicklas Hansen and Hao Su and Xiaolong Wang},
      year={2021},
      eprint={2107.00644},
      archivePrefix={arXiv},
      primaryClass={cs.LG},
      url={https://arxiv.org/abs/2107.00644}, 
}

@misc{yu2021playvirtualaugmentingcycleconsistentvirtual,
      title={PlayVirtual: Augmenting Cycle-Consistent Virtual Trajectories for Reinforcement Learning}, 
      author={Tao Yu and Cuiling Lan and Wenjun Zeng and Mingxiao Feng and Zhizheng Zhang and Zhibo Chen},
      year={2021},
      eprint={2106.04152},
      archivePrefix={arXiv},
      primaryClass={cs.LG},
      url={https://arxiv.org/abs/2106.04152}, 
}

@misc{zheng2024tacotemporallatentactiondriven,
      title={TACO: Temporal Latent Action-Driven Contrastive Loss for Visual Reinforcement Learning}, 
      author={Ruijie Zheng and Xiyao Wang and Yanchao Sun and Shuang Ma and Jieyu Zhao and Huazhe Xu and Hal Daumé III and Furong Huang},
      year={2024},
      eprint={2306.13229},
      archivePrefix={arXiv},
      primaryClass={cs.LG},
      url={https://arxiv.org/abs/2306.13229}, 
}

@misc{grooten2023madilearningmaskdistractions,
      title={MaDi: Learning to Mask Distractions for Generalization in Visual Deep Reinforcement Learning}, 
      author={Bram Grooten and Tristan Tomilin and Gautham Vasan and Matthew E. Taylor and A. Rupam Mahmood and Meng Fang and Mykola Pechenizkiy and Decebal Constantin Mocanu},
      year={2023},
      eprint={2312.15339},
      archivePrefix={arXiv},
      primaryClass={cs.LG},
      url={https://arxiv.org/abs/2312.15339}, 
}

@misc{yu2022maskbasedlatentreconstructionreinforcement,
      title={Mask-based Latent Reconstruction for Reinforcement Learning}, 
      author={Tao Yu and Zhizheng Zhang and Cuiling Lan and Yan Lu and Zhibo Chen},
      year={2022},
      eprint={2201.12096},
      archivePrefix={arXiv},
      primaryClass={cs.LG},
      url={https://arxiv.org/abs/2201.12096}, 
}

@inproceedings{DBLP:conf/cvpr/ChoiLSJSM23,
  author={Hyesong Choi and Hunsang Lee and Wonil Song and Sangryul Jeon and Kwanghoon Sohn and Dongbo Min},
  title={Local-Guided Global: Paired Similarity Representation for Visual Reinforcement Learning},
  year={2023},
  cdate={1672531200000},
  pages={15072-15082},
  url={https://doi.org/10.1109/CVPR52729.2023.01447},
  booktitle={CVPR},
}

@misc{shang2021reinforcementlearninglatentflow,
      title={Reinforcement Learning with Latent Flow}, 
      author={Wenling Shang and Xiaofei Wang and Aravind Srinivas and Aravind Rajeswaran and Yang Gao and Pieter Abbeel and Michael Laskin},
      year={2021},
      eprint={2101.01857},
      archivePrefix={arXiv},
      primaryClass={cs.LG},
      url={https://arxiv.org/abs/2101.01857}, 
}

@article{DBLP:journals/corr/abs-1708-02596,
  author       = {Anusha Nagabandi and
                  Gregory Kahn and
                  Ronald S. Fearing and
                  Sergey Levine},
  title        = {Neural Network Dynamics for Model-Based Deep Reinforcement Learning
                  with Model-Free Fine-Tuning},
  journal      = {CoRR},
  volume       = {abs/1708.02596},
  year         = {2017},
  url          = {http://arxiv.org/abs/1708.02596},
  eprinttype    = {arXiv},
  eprint       = {1708.02596},
  bibsource    = {dblp computer science bibliography, https://dblp.org}
}

@article{DBLP:journals/corr/abs-2010-02193,
  author       = {Danijar Hafner and
                  Timothy P. Lillicrap and
                  Mohammad Norouzi and
                  Jimmy Ba},
  title        = {Mastering Atari with Discrete World Models},
  journal      = {CoRR},
  volume       = {abs/2010.02193},
  year         = {2020},
  url          = {https://arxiv.org/abs/2010.02193},
  eprinttype    = {arXiv},
  eprint       = {2010.02193},
  bibsource    = {dblp computer science bibliography, https://dblp.org}
}

@article{10.1145/122344.122377,
author = {Sutton, Richard S.},
title = {Dyna, an integrated architecture for learning, planning, and reacting},
year = {1991},
issue_date = {Aug. 1991},
publisher = {Association for Computing Machinery},
address = {New York, NY, USA},
volume = {2},
number = {4},
issn = {0163-5719},
url = {https://doi.org/10.1145/122344.122377},
doi = {10.1145/122344.122377},
journal = {SIGART Bull.},
month = jul,
pages = {160–163},
numpages = {4}
}

@misc{MAB,
      title={Why Keep Your Doubts to Yourself? Trading Visual Uncertainties in Multi-Agent Bandit Systems}, 
      author={Jusheng Zhang and Yijia Fan and Kaitong Cai and Jing Yang and Jiawei Yao and Jian Wang and Guanlong Qu and Ziliang Chen and Keze Wang},
      year={2026},
      eprint={2601.18735},
      archivePrefix={arXiv},
      primaryClass={cs.AI},
      url={https://arxiv.org/abs/2601.18735}, 
}

@inproceedings{
KABB,
title={{KABB}: Knowledge-Aware Bayesian Bandits for Dynamic Expert Coordination in Multi-Agent Systems},
author={Jusheng Zhang and Zimeng Huang and Yijia Fan and Ningyuan Liu and Mingyan Li and Zhuojie Yang and Jiawei Yao and Jian Wang and Keze Wang},
booktitle={Forty-second International Conference on Machine Learning},
year={2025},
url={https://openreview.net/forum?id=AKvy9a4jho}
}

@inproceedings{
GAM,
title={{GAM}-Agent: Game-Theoretic and Uncertainty-Aware Collaboration for Complex Visual Reasoning},
author={Jusheng Zhang and Yijia Fan and Wenjun Lin and Ruiqi Chen and Haoyi Jiang and Wenhao Chai and Jian Wang and Keze Wang},
booktitle={The Thirty-ninth Annual Conference on Neural Information Processing Systems},
year={2025},
url={https://openreview.net/forum?id=EKJhU5ioSo}
}

@inproceedings{CF,
  title={{CF}-{VLM}: Counterfactual Vision-Language Fine-tuning},
  author={Jusheng Zhang and Kaitong Cai and Yijia Fan and Jian Wang and Keze Wang},
  booktitle={The Thirty-ninth Annual Conference on Neural Information Processing Systems},
  year={2025},
  url={https://openreview.net/forum?id=0qGtaRTsCo}
}

@inproceedings{
MAT,
title={{MAT}-Agent: Adaptive Multi-Agent Training Optimization},
author={Jusheng Zhang and Kaitong Cai and Yijia Fan and Ningyuan Liu and Keze Wang},
booktitle={The Thirty-ninth Annual Conference on Neural Information Processing Systems},
year={2025},
url={https://openreview.net/forum?id=YDWRTYgR79}
}

@inproceedings{
3DAgent,
title={Tri-{MARF}: A Tri-Modal Multi-Agent Responsive Framework for Comprehensive 3D Object Annotation},
author={Jusheng Zhang and Yijia Fan and Zimo Wen and Jian Wang and Keze Wang},
booktitle={The Thirty-ninth Annual Conference on Neural Information Processing Systems},
year={2025},
url={https://openreview.net/forum?id=YGIbwfNWot}
}

@misc{MMCOT,
      title={MM-CoT:A Benchmark for Probing Visual Chain-of-Thought Reasoning in Multimodal Models}, 
      author={Jusheng Zhang and Kaitong Cai and Xiaoyang Guo and Sidi Liu and Qinhan Lv and Ruiqi Chen and Jing Yang and Yijia Fan and Xiaofei Sun and Jian Wang and Ziliang Chen and Liang Lin and Keze Wang},
      year={2025},
      eprint={2512.08228},
      archivePrefix={arXiv},
      primaryClass={cs.CV},
      url={https://arxiv.org/abs/2512.08228}, 
}

@misc{HTC,
      title={HybridToken-VLM: Hybrid Token Compression for Vision-Language Models}, 
      author={Jusheng Zhang and Xiaoyang Guo and Kaitong Cai and Qinhan Lv and Yijia Fan and Wenhao Chai and Jian Wang and Keze Wang},
      year={2025},
      eprint={2512.08240},
      archivePrefix={arXiv},
      primaryClass={cs.CV},
      url={https://arxiv.org/abs/2512.08240}, 
}

@misc{KAF,
      title={Kolmogorov-Arnold Fourier Networks}, 
      author={Jusheng Zhang and Yijia Fan and Kaitong Cai and Keze Wang},
      year={2025},
      eprint={2502.06018},
      archivePrefix={arXiv},
      primaryClass={cs.LG},
      url={https://arxiv.org/abs/2502.06018}, 
}

@misc{DrDiff,
      title={DrDiff: Dynamic Routing Diffusion with Hierarchical Attention for Breaking the Efficiency-Quality Trade-off}, 
      author={Jusheng Zhang and Yijia Fan and Kaitong Cai and Zimeng Huang and Xiaofei Sun and Jian Wang and Chengpei Tang and Keze Wang},
      year={2025},
      eprint={2509.02785},
      archivePrefix={arXiv},
      primaryClass={cs.CL},
      url={https://arxiv.org/abs/2509.02785}, 
}

@misc{OSC,
      title={OSC: Cognitive Orchestration through Dynamic Knowledge Alignment in Multi-Agent LLM Collaboration}, 
      author={Jusheng Zhang and Yijia Fan and Kaitong Cai and Xiaofei Sun and Keze Wang},
      year={2025},
      eprint={2509.04876},
      archivePrefix={arXiv},
      primaryClass={cs.AI},
      url={https://arxiv.org/abs/2509.04876}, 
}

@misc{VLMDONG,
      title={Learning Dynamics of VLM Finetuning}, 
      author={Jusheng Zhang and Kaitong Cai and Jing Yang and Keze Wang},
      year={2025},
      eprint={2510.11978},
      archivePrefix={arXiv},
      primaryClass={cs.LG},
      url={https://arxiv.org/abs/2510.11978}, 
}

@misc{FDWR,
      title={Failure-Driven Workflow Refinement}, 
      author={Jusheng Zhang and Kaitong Cai and Qinglin Zeng and Ningyuan Liu and Stephen Fan and Ziliang Chen and Keze Wang},
      year={2025},
      eprint={2510.10035},
      archivePrefix={arXiv},
      primaryClass={cs.AI},
      url={https://arxiv.org/abs/2510.10035}, 
}

@misc{RAN,
      title={Rational ANOVA Networks}, 
      author={Jusheng Zhang and Ningyuan Liu and Qinhan Lyu and Jing Yang and Keze Wang},
      year={2026},
      eprint={2602.04006},
      archivePrefix={arXiv},
      primaryClass={cs.LG},
      url={https://arxiv.org/abs/2602.04006}, 
}

@misc{POT,
      title={Process-of-Thought Reasoning for Videos}, 
      author={Jusheng Zhang and Kaitong Cai and Jian Wang and Yongsen Zheng and Kwok-Yan Lam and Keze Wang},
      year={2026},
      eprint={2602.07689},
      archivePrefix={arXiv},
      primaryClass={cs.CV},
      url={https://arxiv.org/abs/2602.07689}, 
}

@misc{SGN,
      title={Spectral Gating Networks}, 
      author={Jusheng Zhang and Yijia Fan and Kaitong Cai and Jing Yang and Yongsen Zheng and Kwok-Yan Lam and Liang Lin and Keze Wang},
      year={2026},
      eprint={2602.07679},
      archivePrefix={arXiv},
      primaryClass={cs.LG},
      url={https://arxiv.org/abs/2602.07679}, 
}

@misc{z13,
      title={Top-Down Semantic Refinement for Image Captioning}, 
      author={Jusheng Zhang and Kaitong Cai and Jing Yang and Jian Wang and Chengpei Tang and Keze Wang},
      year={2025},
      eprint={2510.22391},
      archivePrefix={arXiv},
      primaryClass={cs.CV},
      url={https://arxiv.org/abs/2510.22391}, 
}

@misc{z14,
      title={LLM-CAS: Dynamic Neuron Perturbation for Real-Time Hallucination Correction}, 
      author={Jensen Zhang and Ningyuan Liu and Yijia Fan and Zihao Huang and Qinglin Zeng and Kaitong Cai and Jian Wang and Keze Wang},
      year={2025},
      eprint={2512.18623},
      archivePrefix={arXiv},
      primaryClass={cs.CL},
      url={https://arxiv.org/abs/2512.18623}, 
}

@misc{z15,
      title={DepthSSC: Monocular 3D Semantic Scene Completion via Depth-Spatial Alignment and Voxel Adaptation}, 
      author={Jiawei Yao and Jusheng Zhang and Xiaochao Pan and Tong Wu and Canran Xiao},
      year={2024},
      eprint={2311.17084},
      archivePrefix={arXiv},
      primaryClass={cs.CV},
      url={https://arxiv.org/abs/2311.17084}, 
}

@article{z20,
author = {Li, Xiaohua and Zhang, Jusheng and Safara, Fatemeh},
title = {Improving the Accuracy of Diabetes Diagnosis Applications through a Hybrid Feature Selection Algorithm},
year = {2021},
issue_date = {Feb 2023},
publisher = {Kluwer Academic Publishers},
address = {USA},
volume = {55},
number = {1},
issn = {1370-4621},
url = {https://doi.org/10.1007/s11063-021-10491-0},
doi = {10.1007/s11063-021-10491-0},
journal = {Neural Process. Lett.},
month = mar,
pages = {153–169},
numpages = {17}
}

@misc{z22,
      title={FlashVLM: Text-Guided Visual Token Selection for Large Multimodal Models}, 
      author={Kaitong Cai and Jusheng Zhang and Jing Yang and Yijia Fan and Pengtao Xie and Jian Wang and Keze Wang},
      year={2025},
      eprint={2512.20561},
      archivePrefix={arXiv},
      primaryClass={cs.CV},
      url={https://arxiv.org/abs/2512.20561}, 
}

@inproceedings{M11,
  title={Follow your pose: Pose-guided text-to-video generation using pose-free videos},
  author={Ma, Yue and He, Yingqing and Cun, Xiaodong and Wang, Xintao and Chen, Siran and Li, Xiu and Chen, Qifeng},
  booktitle={Proceedings of the AAAI Conference on Artificial Intelligence},
  volume={38},
  number={5},
  pages={4117--4125},
  year={2024}
}

@inproceedings{M2,
  title={Follow-your-emoji: Fine-controllable and expressive freestyle portrait animation},
  author={Ma, Yue and Liu, Hongyu and Wang, Hongfa and Pan, Heng and He, Yingqing and Yuan, Junkun and Zeng, Ailing and Cai, Chengfei and Shum, Heung-Yeung and Liu, Wei and others},
  booktitle={SIGGRAPH Asia 2024 Conference Papers},
  pages={1--12},
  year={2024}
}

@article{M3,
  title={Follow-your-emoji-faster: Towards efficient, fine-controllable, and expressive freestyle portrait animation},
  author={Ma, Yue and Yan, Zexuan and Liu, Hongyu and Wang, Hongfa and Pan, Heng and He, Yingqing and Yuan, Junkun and Zeng, Ailing and Cai, Chengfei and Shum, Heung-Yeung and others},
  journal={arXiv preprint arXiv:2509.16630},
  year={2025}
}

@article{M4,
  title={Follow-Your-Motion: Video Motion Transfer via Efficient Spatial-Temporal Decoupled Finetuning},
  author={Ma, Yue and Liu, Yulong and Zhu, Qiyuan and Yang, Ayden and Feng, Kunyu and Zhang, Xinhua and Li, Zhifeng and Han, Sirui and Qi, Chenyang and Chen, Qifeng},
  journal={arXiv preprint arXiv:2506.05207},
  year={2025}
}

@article{M5,
  title={Controllable Video Generation: A Survey},
  author={Ma, Yue and Feng, Kunyu and Hu, Zhongyuan and Wang, Xinyu and Wang, Yucheng and Zheng, Mingzhe and He, Xuanhua and Zhu, Chenyang and Liu, Hongyu and He, Yingqing and others},
  journal={arXiv preprint arXiv:2507.16869},
  year={2025}
}

@article{M10,
  title={ContextFlow: Training-Free Video Object Editing via Adaptive Context Enrichment},
  author={Chen, Yiyang and He, Xuanhua and Ma, Xiujun and Ma, Yue},
  journal={arXiv preprint arXiv:2509.17818},
  year={2025}
}

@article{alemi2016deep,
  title={Deep Variational Information Bottleneck},
  author={Alexander A. Alemi and Ian Fischer and Joshua V. Dillon and Kevin Murphy},
  year={2016},
  journal={arXiv preprint arXiv:1612.00410}
}

@article{ba2016layer,
  title={Layer Normalization},
  author={Jimmy Lei Ba and Jamie Ryan Kiros and Geoffrey E. Hinton},
  year={2016},
  journal={arXiv preprint arXiv:1607.06450}
}

@article{ha2018worldmodels,
  title={World Models},
  author={David Ha and J{\"u}rgen Schmidhuber},
  year={2018},
  journal={arXiv preprint arXiv:1803.10122}
}

@misc{hafner2019dreamer,
  title={Dream to Control: Learning Behaviors by Latent Imagination},
  author={Danijar Hafner and Timothy Lillicrap and Ian Fischer and Ruben Villegas and David Ha and Honglak Lee and James Davidson},
  year={2019},
  eprint={1912.01603},
  archivePrefix={arXiv},
  primaryClass={cs.LG}
}

@misc{hafner2019learning,
  title={Learning Latent Dynamics for Planning from Pixels},
  author={Danijar Hafner and Timothy Lillicrap and Ian Fischer and Ruben Villegas and David Ha and Honglak Lee and James Davidson},
  year={2019},
  eprint={1811.04551},
  archivePrefix={arXiv},
  primaryClass={cs.LG}
}

@misc{hafner2020mastering,
  title={Mastering Atari with Discrete World Models},
  author={Danijar Hafner and Timothy Lillicrap and Mohammad Norouzi and Jimmy Ba},
  year={2020},
  eprint={2010.02193},
  archivePrefix={arXiv},
  primaryClass={cs.LG}
}

@article{kingma2013auto,
  title={Auto-Encoding Variational Bayes},
  author={Diederik P. Kingma and Max Welling},
  year={2013},
  journal={arXiv preprint arXiv:1312.6114}
}

@misc{laskin2020rad,
  title={Reinforcement Learning with Augmented Data},
  author={Michael Laskin and Kimin Lee and Adam Stooke and Lerrel Pinto and Pieter Abbeel and Aravind Srinivas},
  year={2020},
  eprint={2004.14990},
  archivePrefix={arXiv},
  primaryClass={cs.LG}
}

@article{mnih2015human,
  title={Human-level control through deep reinforcement learning},
  author={Mnih, Volodymyr and Kavukcuoglu, Koray and Silver, David and Rusu, Andrei A and Veness, Joel and Bellemare, Marc G and Graves, Alex and Riedmiller, Martin and Fidjeland, Andreas K and Ostrovski, Georg and others},
  journal={Nature},
  volume={518},
  number={7540},
  pages={529--533},
  year={2015},
  publisher={Nature Publishing Group}
}

@article{peters2008reinforcement,
  title={Reinforcement learning of motor skills with policy gradients},
  author={Peters, Jan and Schaal, Stefan},
  journal={Neural Networks},
  volume={21},
  number={4},
  pages={682--697},
  year={2008},
  publisher={Elsevier}
}

@article{tassa2018deepmind,
  title={DeepMind Control Suite},
  author={Yuval Tassa and Yotam Doron and Alistair Muldal and Tom Erez and Yazhe Li and Diego de Las Casas and David Budden and Abbas Abdolmaleki and Josh Merel and Andrew Lefrancq and Timothy Lillicrap and Martin Riedmiller},
  year={2018},
  journal={arXiv preprint arXiv:1801.00690}
}

\appendix
\clearpage
\newpage

\end{document}